\pdfoutput=1

\documentclass[11pt]{article}

\usepackage[final]{acl}

\usepackage{times}
\usepackage{latexsym}

\usepackage[T1]{fontenc}

\usepackage[utf8]{inputenc}

\usepackage{microtype}

\usepackage{inconsolata}

\usepackage{graphicx}
\usepackage{multirow}
\usepackage{booktabs}
\usepackage[dvipsnames,table,xcdraw]{xcolor}
\usepackage{bm}
\usepackage{tabularray}
\usepackage{pifont}
\usepackage{stfloats}
\usepackage{float}
\usepackage{caption}
\usepackage{amsmath}
\usepackage{multirow}
\usepackage{booktabs}
\usepackage{arydshln}
\usepackage[normalem]{ulem}
\usepackage{enumitem}
\useunder{\uline}{\ul}{}

\title{Boosting Multi-modal Keyphrase Prediction with Dynamic Chain-of-Thought in Vision-Language Models}

\author{
Qihang Ma, Shengyu Li, Jie Tang, Dingkang Yang, Shaodong Chen, \\
\textbf{Yingyi Zhang, Chao Feng$^\dagger$, Jiao Ran} \\
ByteDance Douyin Content Group \\
\texttt{\{maqihang, lishengyu.129, tangjie.jack, yangdingkang, chenshaodong.rookie\}}, \\ \texttt{\{zhangyingyi.13, chaofeng.zz, ranjiao\}@bytedance.com}
\\ $^{\dagger}$ Corresponding authors
}

\begin{document}

\maketitle

Multi-modal keyphrase prediction (MMKP) aims to advance beyond text-only methods by incorporating multiple modalities of input information to produce a set of conclusive phrases. 
Traditional multi-modal approaches have been proven to have significant limitations in handling the challenging absence and unseen scenarios.
Additionally, we identify shortcomings in existing benchmarks that overestimate model capability due to significant overlap in training tests. 
In this work, we propose leveraging vision-language models (VLMs) for the MMKP task. 
Firstly, we use two widely-used strategies, \textit{e.g.}, zero-shot and supervised fine-tuning (SFT) to assess the lower bound performance of VLMs.
Next, to improve the complex reasoning capabilities of VLMs, we adopt Fine-tune-CoT, which leverages high-quality CoT reasoning data generated by a teacher model to finetune smaller models.  
Finally, to address the “overthinking” phenomenon, we propose a dynamic CoT strategy which adaptively injects CoT data during training, allowing the model to flexibly leverage its reasoning capabilities during the inference stage. 
We evaluate the proposed strategies on various datasets and
the experimental results demonstrate the effectiveness of the proposed approaches. 
The code is available at \textcolor{magenta}{https://github.com/bytedance/DynamicCoT}.
\section{Introduction}

\begin{figure*}[t]
    \centering
    \includegraphics[width=\linewidth]{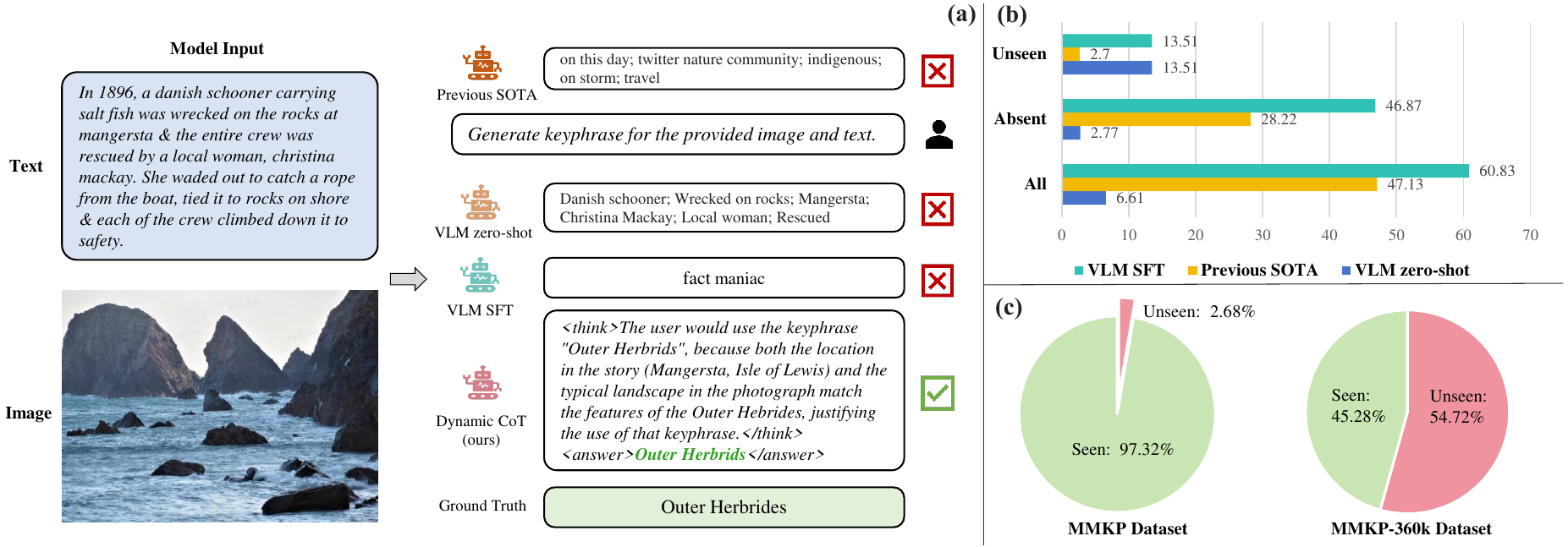}
    \setlength{\abovecaptionskip}{-0.5em}
    \caption{(a) An example of multi-model keyphrase prediction. (b) The performance of different models on the MMKP dataset \citep{wang2020cross}. ``Absent'' refers to keyphrases that absent in the input text. ``Unseen'' refers to keyphrases that not appear in the training set's ground truth. (c) The number of seen and unseen keyphrases in the test set of the MMKP dataset and our MMKP-360k dataset.}
    \label{fig:motivation}
    \vspace{-0.5em}
\end{figure*}

Multi-modal keyphrase prediction (MMKP) aims to generate concise, informative phrases that capture the essence of cross-modal inputs (\textit{e.g.}, text and image inputs in Fig.~\ref{fig:motivation}(a)). Unlike traditional text-only keyphrase prediction \citep{zhang2018encoding, wang2019microblog, zhang2024notellm}, MMKP leverages the complementary nature of visual and textual signals to improve the cross-modal semantic understanding \citep{chang2013towards, bansal2015towards, wang2019topic, zhang2024notellm} and summarization  \citep{davidov2010enhanced, wang2011topic, zhang2024notellm2}. 

Traditional multi-modal approaches~\cite{wang2020cross, dong2023towards} primarily focus on designing cross-modal fusion architectures to integrate visual features (\textit{e.g.}, OCR and visual entities) and textual semantics, by using attention mechanisms or hybrid neural networks to model modality interactions. These methods have achieved great progress in the MMKP task.
 
However, as illustrated in Fig.~\ref{fig:motivation}(b), significant limitations have emerged in more complex scenarios, particularly in handling the following two challenges: 
\begin{itemize}[leftmargin=10pt]
    \item {\em absence scenario}. The case where the prediction keyphrases are lacking in the corresponding input text, requiring the model to exhibit strong cross-modal interaction capabilities and infer keyphrases from cross-modal context.
    \item {\em unseen scenario}: The case where the predicted keyphrases do not appear in the training set, demands that the model possess robust generalization capability. The unseen keyphrases pose a significant challenge to MMKP systems in production environments, where a diverse range of keyphrases emerges daily.
\end{itemize}

In addition, our analysis of the public MMKP~\cite{wang2020cross} dataset and our collected larger-scale (MMKP-360k) production dataset reveal that the two datasets exhibit substantial distribution discrepancies in unseen keyphrase scenarios. The proportion of training samples with unseen keyphrases is only 2.68\% in the public dataset and 54.72\% in the real-word production dataset. The discrepancy in Fig.~\ref{fig:motivation}(b) and Fig.~\ref{fig:motivation}(c), indicates that traditional multi-modal approaches exhibit poor generalization capability in absent and unseen scenarios.

Although VLMs have been widely applied to multi-modal tasks, \textit{e.g.}, visual question answering \citep{antol2015vqa}, image captioning \citep{li2022blip}, video understanding \citep{sigurdsson2016hollywood}, their effectiveness in the MMKP remains underexplored. 
To this end, we propose adopting VLMs for MMKP in an autoregressive manner.
Firstly, we use two widely-used strategies: zero-shot and supervised fine-tuning (SFT). As illustrated in Fig.~\ref{fig:motivation}(b), the SFT approach outperforms the zero-shot approach in the absent scenario but underperforms in the unseen scenario, which indicates that SFT enables the VLMs to leverage its robust vision-language comprehension for similar content, but severely restricts its generalization capability. To solve this, we follow~\citet{ho2023large} to utilize Fine-tune-CoT to improve the complex reasoning capabilities of VLMs. Fine-tune-CoT leverages high-quality CoT reasoning data generated by a teacher model to finetune smaller models. Furthermore, considering the ``overthinking'' phenomenon~\cite{chen2024not} for the seen scenario, we propose a dynamic CoT strategy to enable efficient reasoning~\cite{qu2025survey} for the VLMs. The dynamic CoT strategy enables the VLMs to prefer to choose the non-CoT reasoning for the easy samples (\textit{e.g.}, seen samples).

To ensure the reproducibility of our research, we resampled the public MMKP dataset (MMKP-V2) to match the proportion of seen and unseen parts with that of the MMKP-360k dataset. Furthermore, comprehensive analysis on three datasets confirm that our method significantly improves the generalization capability of VLMs on unseen samples.

The contributions are summarized as follows:
\begin{itemize}
    \item To the best of our knowledge, this work is the first to comprehensively investigate the potential of VLMs for multi-modal keyphrase prediction. 
    \item We propose a Dynamic CoT strategy that enbales VLMs adaptively choosing CoT reasoning ability for the hard unseen samples, which is more suitable in production environments with efficient decoding.
    \item Experimental results and rigorous analysis across multiple datasets validate the efficacy and robustness of our proposed methodology.
\end{itemize}

\section{Related Work}
\label{sec:related_work}
\subsection{Social Media Keyphrase Prediction}
\label{subsec:rela_kp_pred}
Social media keyphrases, including hashtags and categories, serve as concise summaries of user-generated content. 
Prior to the emergence of LLMs, approaches in this domain mainly fell into extractive \citep{zhang2016keyphrase, zhang2018encoding}, classification \citep{zhang2017hashtag, kou2018hashtag, zeng2018topic}, and generative methods \citep{wang2019microblog, wang2019topic, kou2018hashtag}. 
Due to inherent training limitations, the first two types could not produce keyphrases from a truly open set, while generative methods were limited to processing text-only content. With the advent of LLMs, numerous methods \citep{shao2024one2set+, zhang2024notellm, kang2025empirical} have attempted to leverage these models for keyphrases prediction, but most still rely exclusively on textual inputs. 
However, social media posts often contain multi-modal information, thus requiring the model to possess strong multi-modal understanding capabilities.
NoteLLM2 \citep{zhang2024notellm2} uses MLLM and a zero-shot prompt to compress multi-modal posts into a single word vector for end-to-end recommendation model training. However, it does not explore how to generate more comprehensive and accurate keyphrases. In this paper, we investigate the potential of VLMs for keyphrase prediction, fully leveraging multi-modal information to achieve more precise and accurate keyphrase prediction.

\subsection{Vision-Language Models}
\label{subsec:rela_vlm}
Vision-Language Models (VLMs) have emerged as a transformative paradigm in multi-modal learning, bridging visual and textual representations to enable cross-modal understanding and generation \citep{alayrac2022flamingo, li2023blip, openai2023gpt4v, bai2023qwen}. Early efforts in this domain focused on joint embedding spaces \citep{radford2021learning, jia2021scaling}, achieving strong zero-shot transfer capabilities. Subsequent work expanded VLMs to generative tasks, such as image captioning \citep{li2022blip} and visual question answering (VQA) \citep{antol2015vqa}, leveraging architectures like transformer-based encoders-decoders \citep{vaswani2017attention}. Models such as Flamingo \citep{alayrac2022flamingo} and BLIP-2 \citep{li2023blip} integrated pretrained vision encoders with LLMs to unify perception and reasoning, enabling few-shot adaptation to downstream tasks. Later advancements, including GPT-4V \citep{openai2023gpt4v}, Qwen-VL \cite{bai2023qwen} series, Intern-VL \cite{chen2024internvl} series and etc, further scaled training data and model size, demonstrating remarkable performance on complex multi-modal benchmarks.

\subsection{Reasoning Capabilities}
\label{subsec:rela_reason_cap}
Recently, as reasoning models \citep{jaech2024openai, guo2025deepseek} have gained significant attention, inference-time computation has been recognized as an effective approach to further unlock the potential of LLMs. Consequently, an increasing number of studies \citep{xu2024llava, team2025kimi, seed2025seed} have started to incorporate reasoning capabilities into VLMs. In this paper, we examine multiple VLM training paradigms for multi-modal keyphrasse prediction. By integrating world knowledge and reasoning abilities through Dynamic CoT training, our approach enhances model generalization while mitigate the problem of ``overthinking''~\cite{chen2024not}.
\section{Methodology}
\label{sec:method}
In this section, we first discuss paradigms of traditional multi-modal models for MMKP and briefly analyze their limitations (Sec \ref{subsec:task_specific_model}). Next, we focus on how to incorporate reasoning capabilities into VLMs (Sec \ref{subsec:different_vlm_paradigms}), and describe how our proposed approach employs Dynamic CoT to solve the ``overthinking'' phenomenon (Sec \ref{subsec:dynamic_cot}).

\begin{figure*}[ht]
    \centering
    \includegraphics[width=\linewidth]{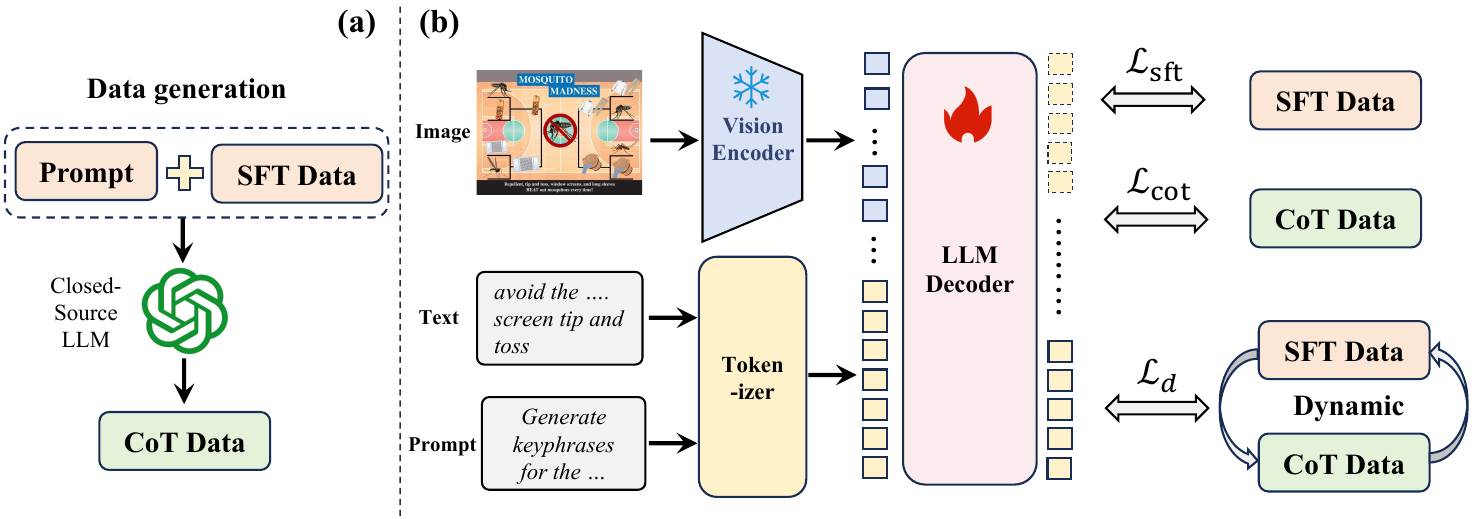}
    \setlength{\abovecaptionskip}{-1.0em}
    \caption{Main framework of our proposed method. (a) CoT data production pipeline. (b) Dynamic CoT training pipeline.}
    \label{fig:main_framework}
    \vspace{-0.5em}
\end{figure*}

\subsection{Traditional Multi-modal Models}
\label{subsec:task_specific_model}
Traditional multi-modal models such as M$^3$H-ATT \citep{wang2020cross} and MM-MKP \citep{dong2023towards} are inherently constrained by limited model capacity and insufficient multi-task capabilities. Consequently, such models typically depend on external Optical Character Recognition (OCR) systems and visual feature extraction modules to augment textual inputs from social media posts. The extracted visual features are concatenated with post text to enhance keyphrase prediction performance. To further improve keyphrase prediction accuracy, these approaches frequently incorporate additional keyphrase classification modules based on all keyphrases in the dataset. The final optimization objective is formulated as a multi-task loss, which is defined as follows:
\begin{equation}
\small
    \mathcal{L}(\theta)=-\sum_{i=1}^{N}[logP_{cls}(\textbf{y}^n)+\gamma \cdot \sum_{t=1}^{l_y^n}logP_{gen}(\textbf{y}_t^{n})],
\end{equation}
where the first term represents the classification loss, while the second term corresponds to the keyword generation loss. Here, $N$ denotes the size of the training set, $\textbf{y}$ is the predicted keyphrase sequence, and $t$ indicates the $t$-th token.

However, this strategy constrains the models' open-set generation capabilities and limits their generalizability. Moreover, the scope of world knowledge embedded in such models remains inadequate. For certain social media posts, such as those involving memes or referencing current events and political topics, a substantial amount of external world knowledge is necessary, which poses significant challenges for traditional multi-modal models.

\subsection{Endow Reasoning Capabilities}
\label{subsec:different_vlm_paradigms}
Compared to traditional multi-modal models, VLMs offer superior capabilities in image-text understanding and generalization. VLMs can effectively comprehend the content of social media posts without relying on external models. For the MMKP task, a straightforward approach to training VLMs involves using the multi-modal content as the input prompt and the ground truth (GT) keyphrases as the response, where the model is optimized using the next-token prediction loss, as shown in Fig. \ref{fig:main_framework}(b). The loss function is defined as follows:
\begin{equation}
\small
    \mathcal{L}_{sft} = -\frac{1}{T} \sum_{t=1}^{T} \log P\left( \textbf{y}^s_t \mid \textbf{y}^s_{<t}, \mathbf{v}; \theta \right),
\end{equation}
here \(\textbf{y}^s = [\textbf{y}_p; \textbf{y}_r^{s}]\), where \(\textbf{y}_p\) and \(\textbf{y}_r^{s}\) denote the input prompt token sequence and the response token sequence, respectively. The response token sequence \(\textbf{y}_r^{s}\) corresponds to the GT keyphrases. \(\mathbf{v}\) denotes the image token sequence, and \(\theta\) denotes the model parameters.

\begin{table}[]
\resizebox{0.48\textwidth}{!}{{\small
\begin{tabular}{lp{0.45\linewidth}}
\toprule
Prompt for Generating CoT data\\ 
\midrule
\begin{tabular}[c]{@{}p{0.95\linewidth}@{}}{[}INST{]}\textless{}SYS\textgreater\\ You are a helpful assistant. Analyze briefly why social media users would use specific hashtags "Keyphrases" for a post titled "Post text" with given image "Image".\textless{}/SYS\textgreater\\ \textless{}USER\textgreater\\ "Keyphrases": \{keyphrases\}\\ "Post text": \{post text\}\\ "Image": \{image\}\\ \textless{}/USER\textgreater{}{[}INST{]}\end{tabular} \\ \bottomrule
\end{tabular}
}}
\caption{The system prompt template for generating CoT responses. ``[INST]'' denotes the instruction provided to the LLM, <SYS> denotes the system prompt and <USER> denotes the user prompt.}
\vspace{-1.0em}
\label{tab:cot_prompt}
\end{table}
Although straightforward SFT demonstrates promising performance on VLMs, it still suffers from several limitations. 
As shown in Fig.~\ref{fig:motivation}(b), the model's generalization is constrained, with its performance on unseen keyphrases nearly matching that of zero-shot scenarios (see Section \ref{subsec:ablation_study} for detailed analysis). Our analysis reveals that supervised fine-tuned (SFT) models exhibit a strong tendency to overfit training samples, generating keyphrases primarily through surface-level pattern matching. Crucially, these models demonstrate limited capability in deciphering the underlying user intent embedded in provided keyphrases, consequently lacking the necessary reasoning capacity to infer contextually appropriate keyphrases.

To overcome these limitations, we constructed multi-modal CoT data, aiming to activate and supplement the model's world knowledge and thereby strengthen its reasoning ability. Specifically, we leveraged GPT-4o \citep{hurst2024gpt} to generate multi-modal CoT data, wherein each thought process centers on analyzing user intent to enhance the model's reasoning capacity while maintaining relevance learning between similar posts. Fig. \ref{fig:main_framework}(a) illustrates the pipeline for constructing CoT data, and the prompt template is shown in Table \ref{tab:cot_prompt}. 

After obtaining such analytic data, the final multi-modal CoT data is organized in the form of ``\texttt{<think>thinking process</think> <answer>keyphrases</answer>}''. The Fine-tune-CoT loss function is formulated as follows:
\begin{equation}
\small
    \mathcal{L}_{cot} = -\frac{1}{T} \sum_{t=1}^{T} \log P\left( \textbf{y}_t^c \mid \textbf{y}^{c}_{<t}, \mathbf{v}; \theta \right),
\end{equation}
where \(\textbf{y}^c = [\textbf{y}_p; \textbf{y}_r^{c}]\) and \(\textbf{y}_r^{c}\) corresponds to the CoT responses.

However, experimental results indicate that the incorporation of CoT data does not yield immediate performance gains. 
Additionally, CoT reasoning introduces extra computational overhead during inference.

\subsection{Dynamic CoT}
\label{subsec:dynamic_cot}
Our analysis reveals two key limitations in Fine-tune-CoT models. First, the "overthinking" phenomenon occurs when the reasoning model generates overly generic keyphrases that fail to capture users' specific preferences. Second, we observe content redundancy in multi-modal CoT generation, where posts sharing identical keyphrases receive highly similar reasoning paths. This redundancy issue becomes particularly severe for high-frequency keyphrases, significantly diminishing model effectiveness.

To more effectively leverage multi-modal CoT data, we propose a dynamic CoT training paradigm, as illustrated in Fig.~\ref{fig:main_framework}(b).

Specifically, during training, we categorize samples as easy or hard based on SFT loss $\mathcal{L}_{sft}$. We hypothesize that overfitting to simple samples during SFT may significantly impair model generalization capability. To mitigate this issue, we introduce a threshold $\gamma$, when the loss of a sample falls below $\gamma$, we switch its supervision to CoT data. The mathematical formula is given as follows:
\begin{equation}
\small
    \mathcal{L}_{d} = -\frac{1}{T} \sum_{t=1}^{T} \log P\left( \textbf{y}_t^d \mid \textbf{y}^{d}_{<t}, \mathbf{v}; \theta \right),
\end{equation}
where
\begin{equation}
\small
    \textbf{y}^d = \left\{
                    \begin{array}{ll}
                       \textbf{y}^c  & \mathcal{L}_{sft} < \gamma\\
                       \textbf{y}^s  & \mathcal{L}_{sft} \ge \gamma\\
                    \end{array}
                    \right. \\.
\end{equation}

This dynamic adjustment of the supervision signal enables the model to adapt its output format based on input complexity, thereby further enhancing generalization while maintaining robust relevance learning.

\section{Experiments}
\label{sec:experiments}

\begin{table}[t]
\setlength{\tabcolsep}{0.015\linewidth}
\begin{center}
\resizebox{0.45\textwidth}{!}{{\small
    \begin{tabular}{lcccc}
    \toprule
    \multicolumn{1}{c}{\multirow{2}{*}{Datasets}} & \# Train & \# Test & \# KP  & Train~$|$KP$|$        \\
    \multicolumn{1}{c}{}                          & Posts    & Posts   & / Post & $\cap$ Test~$|$KP$|$         \\
    \midrule
    MMKP    & 42,959 & 5,372 & 1.33 & 97.32\%                              \\
    MMKP-V2 & 34,515 & 10,564 & 1.29 & 44.92\%                            \\
    MMKP-360k & 330,614 & 36,736 & 4.48 & 45.28\%                            \\ \bottomrule
    \end{tabular}
}}
\end{center}
\caption{Statistics of different datasets. KP: keyphrase; Train $|$KP$|$: the size of unique keyphrase in train set. $\cap$ denotes the intersection of the two sets.}
\label{tab:dataset_statics}
\end{table}
\begin{figure}[ht]
    \centering
    \includegraphics[width=0.8\linewidth]{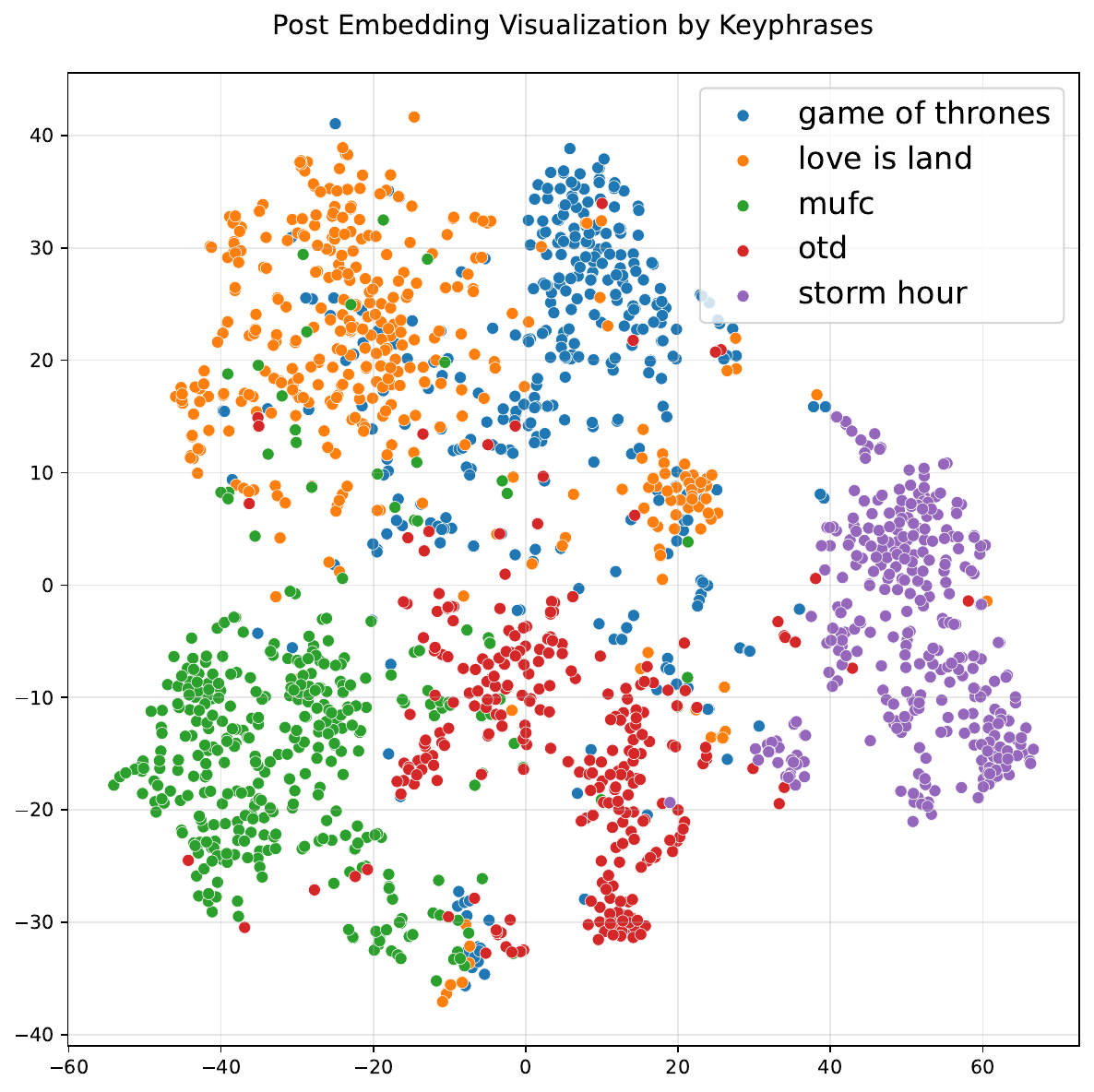}
    \caption{Visualization of multi-modal embedding clustering for post sharing the same keyphrase (The top five most frequent keyphrases) in the MMKP dataset.}
    \label{fig:kp_tsne}
\end{figure}

\subsection{Datasets}
\label{subsec:datasets}
\textbf{MMKP Dataset.}
Multi-modal Keyphrase Prediction (MMKP) Dataset was collected by \citep{wang2020cross}. This dataset includes 53,701 English samples, each of which comprises a distinct text-image pair, with user-annotated hashtags serving as keyphrases. The dataset has a diverse set of categories and only around half of the images (54\%) are natural photos, which is rather different from other standard image data such as MS-COCO \citep{lin2014microsoft}. Furthermore, 52\% of the samples contain either semantically uninformative text or irrelevant images, indicating a complex and often discordant multimodal relationship. The authors randomly split the data into 80\%, 10\%, 10\% corresponding to training, validation, and test set. As shown in Table~\ref{tab:dataset_statics}, there are 4,261 unique keyphrases in the training set and 2,534 unique keyphrases in the test set, among which 2,466 keyphrases from the test set also appear in the training set, resulting in a high overlap rate of 97.32\%.

\begin{table*}[ht]
\setlength{\tabcolsep}{0.009\linewidth}
\begin{center}
\resizebox{\textwidth}{!}{{\small
    \begin{tabular}{ccccccccccc}
    \toprule
    \multirow{2}{*}{\bf Models}  & \multirow{2}{*}{\bf Training}     & \bf MMKP & \multicolumn{3}{c}{\bf MMKP-V2} & \multicolumn{3}{c}{\bf MMKP-360k} & {\bf Avg} \\ \cline{3-11} 
                                  & & All & All    & Absent    & Unseen   & All     & Absent    & Unseen    & All \\ \midrule
    \multicolumn{10}{c}{Baseline Models}                                           \\
    \midrule
    \multicolumn{2}{c}{CO-ATT \citep{zhang2017hashtag}} & 42.12 & - & - & - & - & - & - & -           \\
     \multicolumn{2}{c}{M$^3$H-ATT \citep{wang2020cross}}  & 47.06 & - & - & - & - & - & - & -            \\
     \multicolumn{2}{c}{MM-MKP \citep{dong2023towards}} & 48.19 & - & - & - & - & - & - & -             \\
    \midrule
    \multicolumn{10}{c}{Our Experiments}           \\
    \midrule
    \multirow{3}{*}{InternVL3-2B} & zero-shot & 3.62 & 4.56 & 1.39 & 4.53 & 19.00 & 5.64 & 9.62 & 9.06 \\
    & full sft & 57.54 & 28.89 & 18.10 & 7.98 & 38.53 & 18.47 & 24.42 & 41.65\\\rowcolor[HTML]{EFEFEF}
    & Dynamic CoT & \textbf{59.63} & \textbf{30.76} & \textbf{18.99} & \textbf{9.90} & \textbf{40.03} & \textbf{19.95} & \textbf{26.04} & \textbf{43.47}\\
    \hdashline
    \multirow{3}{*}{InternVL3-8B} & zero-shot & 6.26 & 6.89 & 2.37 & 6.72 & 11.43 & 3.26 & 7.05 & 8.19  \\
    & full sft & 57.83 & 28.58 & 18.30 & 7.17 & 40.48 & 19.88 & 25.45 & 42.30 \\\rowcolor[HTML]{EFEFEF}
    & Dynamic CoT & \textbf{60.29} & \textbf{31.42} & \textbf{19.13} & \textbf{10.68} & \textbf{50.53} & \textbf{20.04} & \textbf{26.44} &  \textbf{47.41}\\
    \hdashline
     \multirow{3}{*}{Qwen2.5-VL-3B} & zero-shot & 4.48 & 4.50 & 1.50 & 4.37 & 17.08 & 4.93 & 11.29 & 8.49\\
    & full sft & 60.33 & 29.89 & 19.59 & 8.79 & 43.04 & 22.06 & 24.60 & 44.42 \\\rowcolor[HTML]{EFEFEF}
    & Dynamic CoT & \textbf{61.90} & \textbf{33.14} & \textbf{20.19} & \textbf{12.48} & \textbf{47.51} & \textbf{22.62} & \textbf{26.68} & \textbf{47.52} \\
    \hdashline
    \multirow{3}{*}{Qwen2.5-VL-7B} & zero-shot & 6.61 & 7.75 & 2.75 & 8.38 & 14.34 & 4.10 & 9.94 & 9.57  \\
      & full sft & 60.83 & 30.49 & 20.90 & 7.90 & 43.70 & 22.28 & 24.98 & 45.01  \\\rowcolor[HTML]{EFEFEF}
      & Dynamic CoT & \textbf{63.58} & \textbf{33.56} & \textbf{22.32} & \textbf{13.36} & \textbf{50.66} & \textbf{23.41} & \textbf{26.43} & \textbf{49.27} \\
    \bottomrule
    \end{tabular}
}}
\end{center}
\caption{Performance comparison for multi-modal keyphrase prediction task. We adopt F1@1 (\%) as the evaluation metric for the MMKP and MMKP-V2 datasets, while F1@M (\%) is employed for the MMKP-360k Dataset.}
\vspace{-0.2em}
\label{tab:sota_compare}
\end{table*}

\textbf{MMKP-360k Dataset.}
To better evaluate the effectiveness of our proposed method, we constructed a larger-scale multi-modal keyphrase prediction dataset collected based on user-generated contents publicly available on internet platforms. Following the construction methodology of the MMKP dataset, 
we extracted the users’ hashtag data, which was subsequently cleaned and refined using LLMs. The processed hashtags served as the final ground truth keyphrases. The resulting MMKP-360k Dataset comprises 330,614 training samples and 36,736 test samples, each of which comprises a distinct text-image pair. There are 502k unique keyphrases in the training set and 81k unique keyphrases in the test set, of which 37k keyphrases from the test set also occur in the training set, resulting in an overlap rate of 45.28\%.

\textbf{Resampled MMKP Dataset (MMKP-V2).}
According to the keyphrase statistics for the MMKP dataset, the vast majority of keyphrases in the test set also appear in the training set. This enables models to significantly improve accuracy on the test set by simply fitting the training data and learning the similarity among posts containing identical keyphrases. However, statistics from our collected MMKP-360K data indicate that the overlap rate is only 45.28\%, and as time progresses, an increasing number of new keyphrases are being created by users. This suggests that models require stronger generalization and reasoning capabilities in order to more accurately predict keyphrases for posts that express previously unseen main ideas.

To enhance alignment with real-world data distributions, we reconstructed the MMKP dataset through two key modifications: 1. Transferring all training-exclusive keyphrase samples to the test set. 2. Removing test samples containing keyphrases observed during training. The resulting MMKP-V2 dataset contains 34,515 samples in the training set and 10,564 samples in the test set, with 2,455 and 3,297 unique keyphrases respectively. Notably, 1,481 test keyphrases from the test set also appear in the training set, resulting in an overlap rate of 44.92\%.

\subsection{Experimental Setup}
\label{subsec:experimental_setup}
To ensure a fair comparison, all models were trained with identical hyperparameter configurations on both the MMKP and MMKP-V2 datasets. Specifically, we employed the AdamW optimizer with an initial learning rate of $5 \times 10^{-5}$, using a cosine annealing schedule for learning rate adjustment. During the SFT process, the parameters of the visual module were frozen, and only the multi-modal projector as well as the large language model components were fine-tuned. The batch size was consistently set to 1 across all experiments. Models with 2B or 3B parameters were trained for 5 epochs, while those with larger parameter sizes were uniformly trained for 3 epochs. Given that the average number of keyphrases per post is approximately 1.3, F1@1 is adopted as evaluation metric. We utilized the GPT-4o-2024-05-13 \citep{hurst2024gpt} to generate CoT reasoning data. The Dynamic CoT loss threshold $\gamma$ is set to $0.4$ for all the models. For the MMKP-360k dataset, we employed the AdamW optimizer with an initial learning rate of $3 \times 10^{-5}$. All models were trained for 3 epochs. We utilized the Doubao-1.5-pro \citep{seed2025doubaov15prp} to generate CoT reasoning data. We adopt F1@M as evaluation metrics, where M denotes the number of keyphrases predicted by the model. 

To validate the universality of our approach, we performed experiments across multiple LLMs and VLMs, including Llama-3.2 \citep{Meta2024Llama32}, Qwen2.5 \citep{yang2024qwen2}, Llama-3.2-Vision \citep{Meta2024Llama32}, InternVL-3 \citep{zhu2025internvl3} and Qwen2.5-VL \citep{bai2025qwen2}.

\subsection{Comparing with SOTA methods}
\label{subsec:compare_sota}
Table~\ref{tab:sota_compare} compares the performance of our approach with baseline models. The table is organized vertically into two categories: baseline traditional multi-modal models and VLMs. Horizontally according to results on the MMKP, MMKP-V2 and MMKP-360k datasets. Detailed experimental results for additional text-only models can be found in the Appendix.

As shown in Table~\ref{tab:sota_compare}, the SFT VLMs outperform state-of-the-art multi-modal keyphrase prediction models (\textit{e.g.}, M$^3$H-ATT\citep{wang2020cross} and MM-MKP \citep{dong2023towards}) by over 20\%. These results suggest that VLMs, which possess broader world knowledge, offer a higher upper bound for multi-modal keyphrase prediction tasks and are a preferable choice for such applications. Furthermore, our method achieves consistent improvements over zero-shot and SFT across various datasets and baselines, and shows significant gains with respect to unseen keyphrases, highlighting the robustness and generalizability of our approach.

\begin{table*}[ht]
\setlength{\tabcolsep}{0.010\linewidth}
\begin{center}
\resizebox{\textwidth}{!}{{\small
    \begin{tabular}{ccccccccccc}
    \toprule
    \multirow{2}{*}{\bf Models} & \multicolumn{3}{c}{\bf MMKP} & \multicolumn{3}{c}{\bf MMKP-V2} & \multicolumn{4}{c}{\bf Avg}    \\ \cline{2-11} 
                  & All    & Seen   & Unseen  & All      & Seen     & Unseen    & All & $\Delta$ (\%) & Unseen & $\Delta$ (\%) \\ \midrule
    Qwen2.5-VL-3B & 4.48 & 5.26 & 12.16 & 4.50 & 4.42 & 4.37 & 4.49 &    - & 8.27 & - \\ \midrule
    + SFT         & 60.33  & 61.26  & 12.16   & 29.89    & 55.88    & 8.79      & 45.11 & - & 10.48 & -\\
    + Fine-tune-CoT & 56.99  & 57.94  & 9.46    & 31.88    & 53.79    & \textbf{13.57}     & 44.44 & {\color{red} $\downarrow$ 1.49} &    11.52 & {\color{blue} $\uparrow$ 9.92}     \\
    + Multi-task  & 60.87  & 61.75  & 9.46    & 31.53    & \textbf{57.96}    & 10.26     & 46.20 & {\color{blue} $\uparrow$ 2.42} & 9.86 & {\color{red} $\downarrow$ 5.92} \\
    \rowcolor[HTML]{EFEFEF}
    + Dynamic CoT       & \textbf{61.27} & \textbf{61.83} & \textbf{14.87} & \textbf{33.14} & 57.14 & 12.48 & \textbf{47.21} &{\color{blue} $\uparrow$ 4.66} & \textbf{13.68} & {\color{blue} $\uparrow$ 30.53}\\ \midrule
    Qwen2.5-VL-7B & 6.61   & 8.70   & 13.51   & 7.75 & 11.22 & 8.38 & 7.18 & - & 10.95 & -     \\ \midrule
    + SFT         & 60.83  & 61.60  & \textbf{13.51}   & 30.49 & 58.45 & 7.90 & 45.66 & - & 10.71 & - \\
    + Fine-tune-CoT  & 61.97  & 62.55  & 12.16   & 33.53 & 57.46 & \textbf{13.42} & 47.75 & {\color{blue} $\uparrow$ 4.58} & 12.79 & {\color{blue} $\uparrow$ 19.42} \\
    + Multi-task  & 62.29  & 63.09  & \textbf{13.51}   & 31.87 & \textbf{59.74} & 9.48 & 47.08 & {\color{blue} $\uparrow$ 3.11} & 11.50 & {\color{blue} $\uparrow$ 7.38}     \\
    \rowcolor[HTML]{EFEFEF}
    + Dynamic CoT & \textbf{63.58}  & \textbf{64.22}  & \textbf{13.51} & \textbf{33.56} & 58.56 & 12.24 & \textbf{48.57} & {\color{blue} $\uparrow$ 6.37} & \textbf{12.89} & {\color{blue} $\uparrow$ 20.35} \\ \bottomrule
    \end{tabular}
}}
\end{center}
\caption{Performance comparison for different training strategy. In the multi-task setting, we treat CoT data as an additional training objective, while keeping the number of training steps consistent with other methods.}
\label{tab:ablation_strategy}
\end{table*}
\subsection{Ablation Study}
\label{subsec:ablation_study}
In this section, we present extensive ablation studies with detailed results summarized in Table~\ref{tab:ablation_strategy}. We provide a thorough analysis of the findings and demonstrate that our proposed method effectively enhances the generalization capability of the model.

\begin{figure*}[ht]
    \centering
    \includegraphics[width=\linewidth]{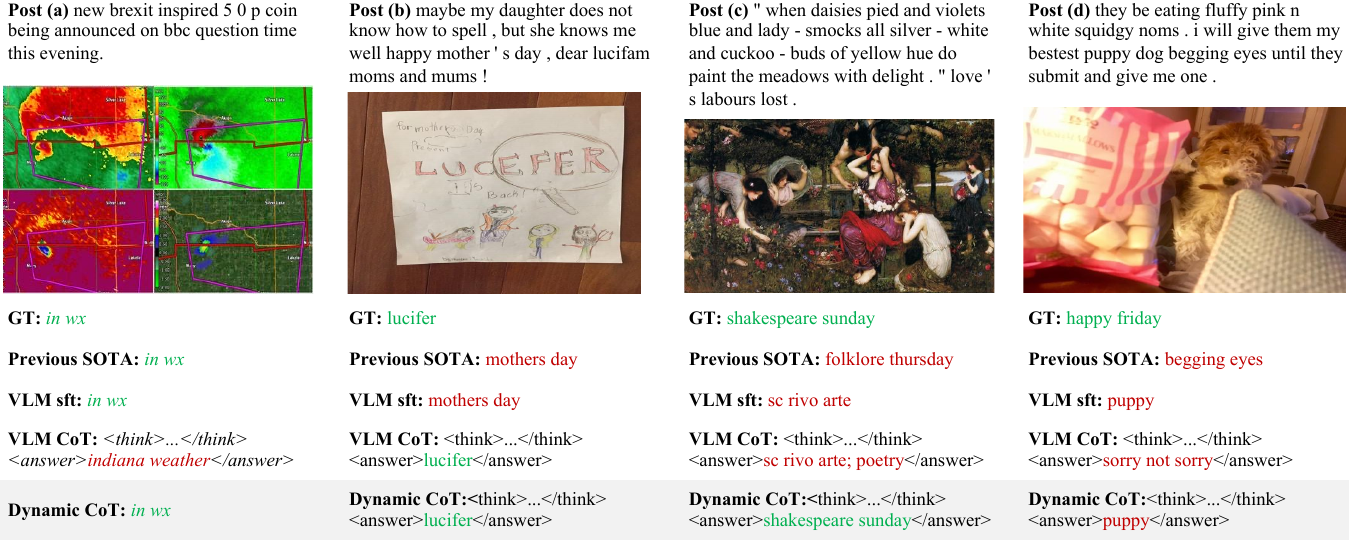}
    \setlength{\abovecaptionskip}{-0.5em}
    \caption{Examples of Multi-modal Keyphrase Prediction. Green denotes correct keyphrase predictions, whereas red denotes incorrect keyphrase predictions.}
    \label{fig:case_study}
\end{figure*}

\textbf{Why is simple SFT so effective, while Fine-tune-CoT leads to a drop in performance?} We believe there are two main reasons for this. 
First, in MMKP task, \textit{effectively learning inter-posts relevance plays a critical role}. This task to some extent reflects the commonalities of group behavior, users tend to select hashtags that resonate with the group to help the post spread better within the community. By analyzing training set posts that share the same keyphrase as shown in Fig.~\ref{fig:kp_tsne}, we observe that their content often exhibits notable similarity. Furthermore, many of these keyphrases serve as abstractions that encapsulate the essence of such posts, rather than functioning as straightforward, literal summaries. Additionally, \textit{Fine-tune-CoT model has “overthinking” phenomenon}. 
Although Fine-tune-CoT model is able to infer keyphrases that semantically correspond to the post content, the generated keyphrases tend not to match those commonly used by social media users, deviating from typical user preferences, for example, generating "indiana weather" instead of the more user-preferred "in wx", as shown in Fig.~\ref{fig:case_study} Post (a).

Second, \textit{significant training-test overlap overestimate the model's capability}. According to the previous analysis, keyphrases in the MMKP dataset have a high overlap between the training and test sets. The model can achieve excellent performance on the test set by taking the shortcut of fitting the training samples. As shown in the Table~\ref{tab:ablation_strategy}, the improvement in test set performance mainly comes from seen keyphrases, while the performance on unseen keyphrases is basically similar to zero-shot. Additionally, by observing the SFT model's case performance on the test set, we found that the model generates completely unrelated KP words for some posts, and the content of these posts is similar to those in the training set that contain incorrectly predicted keyphrases. \textit{This means that the SFT model is relying more on similar memory for keyphrases prediction, rather than reasoning ability.}

\textbf{Balancing commonality and generalization.}
Based on the preceding analysis, we recognize that to effectively address MMKP task, a model must develop dual capabilities: leveraging commonality for prediction with seen keyphrases while employing generalization for unseen keyphrases.
Our proposed method dynamically regulates chain-of-thought learning according to the difficulty of the samples, leading to a better trade-off between commonality learning and generalization ability. As shown in Fig.~\ref{fig:case_study}, during inference, the model adaptively selects either direct keyphrase prediction or prediction with CoT based on individual samples. This approach enhances model accuracy while simultaneously maintaining a balance with inference costs. As a result, our approach achieves optimal performance, with improvements on unseen samples reaching up to 20-30\%.

\textbf{Ablation study on the dynamic CoT loss threshold $\gamma$.}
The $\gamma$ parameter serves as a critical factor in our experiments, as it defines the boundary between SFT and CoT data. We conducted a rigorous ablation study to investigate the impact of various thresholds, as detailed in Table~\ref{tab:gamma ablation}. The initial threshold was selected based on the convergence behavior of the model’s loss curve during the SFT stage, and an adaptive threshold was also evaluated. Experimental results indicate that all tested thresholds yield improvements over the baseline, with $\gamma=0.4$ achieving the best performance. The adaptive threshold, however, did not reach optimal results in our experiments.
\begin{table}[]
\begin{center}
\setlength{\tabcolsep}{0.05\linewidth}
\resizebox{0.45\textwidth}{!}{{\small
\begin{tabular}{cccc}
\toprule
\multirow{2}{*}{\bf $\gamma$} & \multicolumn{3}{c}{\bf MMKP-V2} \\ \cline{2-4} 
                       & All     & Absent   & Unseen  \\ \midrule
baseline               & 29.89   & 19.59 & 8.79    \\
avg                    & 32.22   & 19.50 & \textbf{13.26}   \\
0.3                    & 33.07 & \textbf{20.51} & 12.14 \\
0.4                    & \textbf{33.14}   & 20.19    & 12.48   \\
0.5                    & 32.65   & 19.91    & 12.26   \\
0.6                    & 32.59   & 19.65    & 12.40   \\ \bottomrule
\end{tabular}
}}
\end{center}
\setlength{\abovecaptionskip}{-0.1em}
\caption{Ablation study on Dynamic CoT threshold $\gamma$. ``baseline'' denotes Qwen2.5-VL-3B. ``avg'' indicates that the average train loss is used as a dynamic threshold during training.}
\vspace{-0.8em}
\label{tab:gamma ablation}
\end{table}
\section{Conclusion}
\label{sec:conclusion}
This study provides a comprehensive investigation for employing Vision-Language Models (VLMs) in multi-modal keyphrase prediction. To enhance model generalization, we developed multi-modal CoT data that advance VLMs' reasoning capabilities. Considering the ``overthinking'' phenomenon, we propose a Dynamic CoT training strategy that adaptively optimizes reasoning processes while preserving model generalizability and computational efficiency. Furthermore, by constructing new datasets with reduced train-test keyphrase overlap, we provide a more realistic evaluation of model generalization. Experimental results confirm that VLMs equipped with Dynamic CoT significantly outperform traditional multi-modal approaches, demonstrating superior cross-modal understanding and enhanced generalization.

\nocite{Ando2005,andrew2007scalable,rasooli-tetrault-2015}

\section*{Acknowledgements}
We sincerely appreciate our colleagues at ByteDance for their support. They contributed to this work but were not listed as authors. All contributors are listed in alphabetical order by last name: Sen Cheng, Jing Liang, Xinning Wang, Xinyi Wang, Ting Wen, Danhe Yang.

\section*{Limitations}

While our framework demonstrates promising results on Multi-modal Keyphrase Prediction (MMKP) task through Vision-Language Models (VLMs) and Dynamic Chain-of-Thought (Dynamic CoT), three primary limitations persist. First, the threshold determination in Dynamic CoT remains empirically driven. Despite testing adaptive threshold selection strategies, we observed suboptimal performance in dynamic adjustment. Second, VLMs inherently possess substantially larger parameter counts (e.g., 2B+) compared to traditional multi-modal models, resulting in elevated computational overhead during inference. In addition, incorporating reasoning capabilities further scales the test-time computation.
Thrid, the cost of generating CoT data is relatively high, which poses challenges for the creation of larger-scale CoT datasets.

\section*{Ethics Statement}
All the data utilized in our study was sourced from publicly available content on internet platforms. The seed instructions, which are openly accessible, comply with their respective open-source licenses. Furthermore, these datasets exclude any instances that could give rise to ethical concerns, such as unauthorized sensitive information, thereby minimizing potential societal risks.

\bibliography{custom}
\clearpage
\appendix
\section{Appendix}
\label{sec:appendix}

\begin{table*}[ht]
\setlength{\tabcolsep}{0.01\linewidth}
\setlength{\abovecaptionskip}{-0.4em}
\begin{center}
    \begin{tabular}{ccccccccc}
    \toprule
    \multirow{2}{*}{\bf Models}       & {\bf MMKP} & \multicolumn{3}{c}{\bf MMKP-V2} & \multicolumn{3}{c}{\bf MMKP-360k} & {\bf Avg} \\ \cline{2-9} 
                                  & All & All    & Absent    & Unseen   & All     & Absent    & Unseen    & All \\ \midrule
    \multicolumn{9}{c}{Image-only models}                                                                                      \\ \midrule
    VGG                           & 15.69    & - & - & - & - & - & - & -          \\
    BUTD \citep{anderson2018bottom} & 20.02    & - & - & - & - & - & - & -            \\ \midrule
    \multicolumn{9}{c}{Text-only models}                                                                                       \\ \midrule
    ONE2SEQ \citep{yuan2018one}   & 38.05    & - & - & - & - & - & - & -           \\
    ONE2SET \citep{ye2021one2set} & 36.36    & - & - & - & - & - & - & -             \\
    TOPIC \citep{wang2019topic}                         & 43.17    & - & - & - & - & - & - & -         \\
    Llama-3.2-2B         & 43.50 & 21.92 & 8.62 & 6.28 & 36.49 & 19.22 & 25.98 &  33.97           \\
    Qwen2.5-3B           & 48.53 & 24.33 & 12.35 & 6.58 & 37.21 & 19.99 & 26.04 &  36.69            \\
    Qwen2.5-7B           & 48.84 & 22.74 & 12.40 & 3.94 & 37.75 & 20.62 & 26.68 &  36.44            \\ \midrule
    \multicolumn{9}{c}{Image-text models}                                                                                      \\ \midrule
    CO-ATT \citep{zhang2017hashtag} & 42.12 & - & - & - & - & - & - & -           \\
    M$^3$H-ATT \citep{wang2020cross}  & 47.06 & - & - & - & - & - & - & -           \\
    MM-MKP \citep{dong2023towards} & 48.19 & - & - & - & - & - & - & -             \\
    Llama-3.2-11B-Vision & 59.81 & 28.30 & 20.92 & 5.45 & - & - & - & -                   \\ \midrule
    InternVL-3-2B & 57.54 & 28.89 & 18.10 & 7.98 & 38.53 & 18.47 & 24.42 & 41.65           \\
    \rowcolor[HTML]{EFEFEF}
    + Dynamic CoT (ours) & \textbf{59.63} & \textbf{30.76} & \textbf{18.99} & \textbf{9.90} & \textbf{40.03} & \textbf{19.95} & \textbf{26.04} & \textbf{43.47} \\ \midrule
    InternVL-3-8B & 57.83 & 28.58 & 18.30 & 7.17 & 40.48 & 19.88 & 25.45 & 42.30 \\
    \rowcolor[HTML]{EFEFEF}
    +Dynamic CoT (ours) &\textbf{60.29} & \textbf{31.42} & \textbf{19.13} & \textbf{10.68} & \textbf{50.53} & \textbf{20.04} & \textbf{26.44} &  \textbf{47.41} \\ \midrule
    Qwen2.5-VL-3B & 60.33 & 29.89 & 19.59 & 8.79 & 43.04 & 22.06 & 24.60 & 44.42         \\
    \rowcolor[HTML]{EFEFEF}
    + Dynamic CoT (ours) & \textbf{61.90} & \textbf{33.14} & \textbf{20.19} & \textbf{12.48} & \textbf{47.51} & \textbf{22.62} & \textbf{26.68} & \textbf{47.52}\\ \midrule
    Qwen2.5-VL-7B & 60.83 & 30.49 & 20.90 & 7.90 & 43.70 & 22.28 & 24.98 & 45.01 \\
    \rowcolor[HTML]{EFEFEF}
    +Dynamic CoT (ours) & \textbf{63.58} & \textbf{33.56} & \textbf{22.32} & \textbf{13.36} & \textbf{50.66} & \textbf{23.41} & \textbf{26.43} & \textbf{49.27}\\ \bottomrule
    \end{tabular}
\end{center}
\caption{Performance comparison for multi-modal keyphrase prediction task.}
\vspace{-0.5em}
\label{tab:appendix_sota}
\end{table*}

\textbf{Challenges faced by SOTA MLLMs.}
As discussed in Section~\ref{subsec:ablation_study}, we focus on keyphrase generation for social media posts, which substantially differs from keyword generation for academic papers or news. Social media users typically assign hashtags to their posts not just to summarize the content, but to attract specific audiences, follow trending topics, or reflect elements of community culture such as memes (as shown in Fig.~\ref{fig:case_study} Post (b)). To address these unique characteristics, the model must not only comprehend the content of the post itself but also infer the underlying intent\cite{chang2023latent} of the user in order to generate more suitable keyphrases. This remains difficult even for SOTA MLLMs when they have not been trained on task-specific datasets.

\textbf{Comprehensive Experimental Results.}
Table~\ref{tab:appendix_sota} presents detailed experimental results for additional model variants, including the image-only and text-only models.

multi-modal models demonstrate superior performance over pure visual or pure language models on MMKP task involving multi-modal social media posts. Extracting key information from such posts typically requires integrating both visual and textual cues.

\begin{table}[t]
\centering
\begin{tabular}{cc}
\hline
\textbf{Models} & \textbf{Output Length} \\ \hline
Ground Truth    & 1.33                   \\ \hline
+SFT            & 1.89                   \\
+Fine-tune-CoT  & 157.37                 \\
+Dynamic CoT    & 96.82                  \\ \hline
\end{tabular}
\caption{Inference sequence lengths (measured in words) of different model variants on the MMKP-V2 dataset, all implemented using the Qwen2.5-VL-3B.}
\label{tab:infer_len}
\end{table}
\textbf{Inference Length.}
Table~\ref{tab:infer_len} presents the inference sequence lengths across model variants, demonstrating that our dynamic Chain-of-Thought (CoT) strategy substantially reduces CoT reasoning steps while decreasing computational overhead by 38.48\%.

\begin{figure*}[ht]
    \centering
    \includegraphics[width=0.8\linewidth]{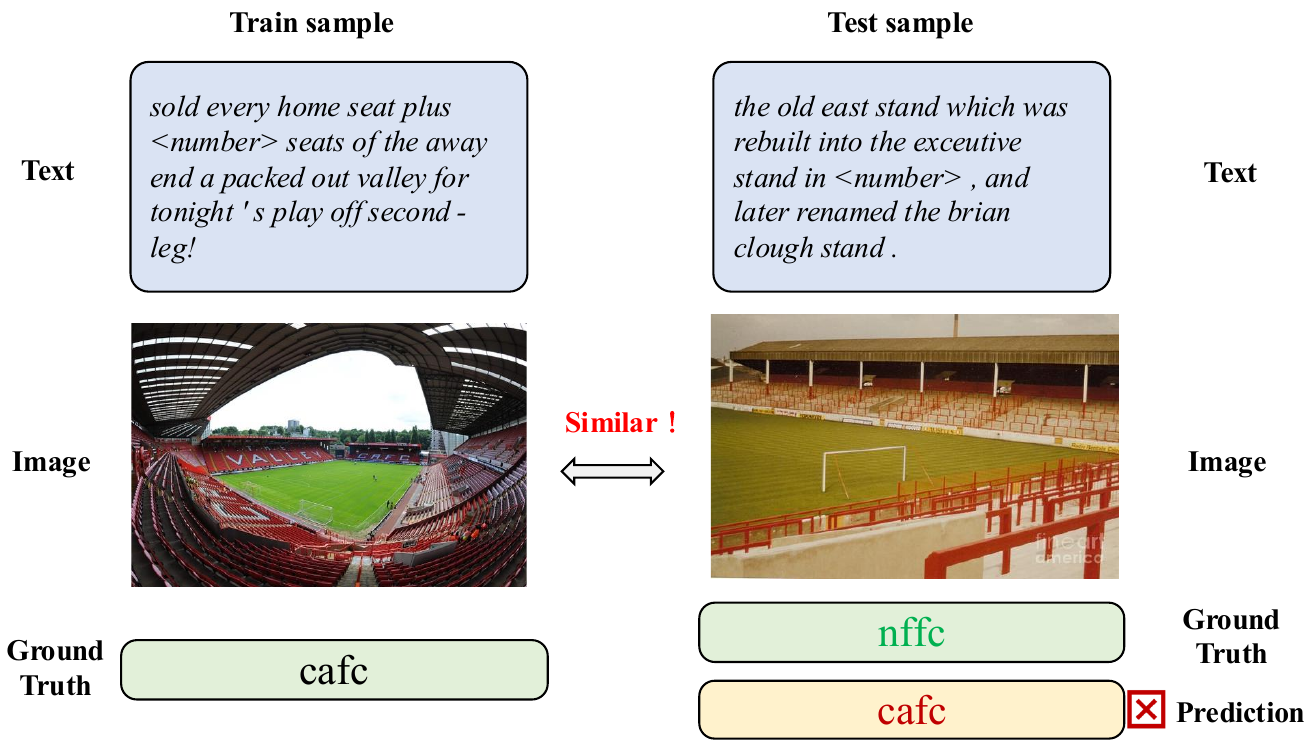}
    \caption{Visualization of SFT models on test set of MMKP dataset.}
    \label{fig:overfit}
\end{figure*}
\textbf{The Overfitting Phenomenon in SFT Models.}

\begin{table*}[ht]
\setlength{\tabcolsep}{0.010\linewidth}
\begin{center}
\resizebox{\textwidth}{!}{{\small
    \begin{tabular}{cccccccc}
    \toprule
    \multirow{2}{*}{\bf Models} & \multirow{2}{*}{\bf Training} & \multicolumn{3}{c}{\bf MMKP} & \multicolumn{3}{c}{\bf MMKP-V2} \\ \cline{3-8} 
                                &                               & F1 $\uparrow$   & GPT4o $\downarrow$   & Human $\downarrow$ & F1 $\uparrow$  & GPT4o $\downarrow$   & Human $\downarrow$ \\ \midrule
    MM-MKP & SFT & 48.19 & 2.50 & 2.23 & - & 2.96 & 2.30 \\ \midrule
    GPT 4o-0513 & zero-shot & 3.71 & 2.80 & 1.95 & 4.86 & 2.94 & 2.10 \\
    Qwen2.5-VL-72B & zero-shot & 5.91 & 2.98 & 2.08 & 7.48 & 3.40 & 2.42 \\
    Qwen2.5-VL-3B & SFT & 60.33 & 1.96 & 1.65 & 29.89 & 2.22 & 1.74 \\
    Qwen2.5-VL-3B & Fine-tune-CoT & 56.99 & 2.10 & 1.85 & 31.88 & 2.02 & 1.64\\
    \rowcolor[HTML]{EFEFEF}
    Qwen2.5-VL-3B & Dynamic CoT & 61.27 & 1.94 & 1.63 & 33.14 & 1.76 & 1.46 \\ \bottomrule
    \end{tabular}
}}
\end{center}
\caption{Performance comparison for different metrics.}
\label{tab:appendix_compare_f1_human}
\end{table*}

\begin{figure*}[!ht]
    \centering
    \includegraphics[width=\linewidth]{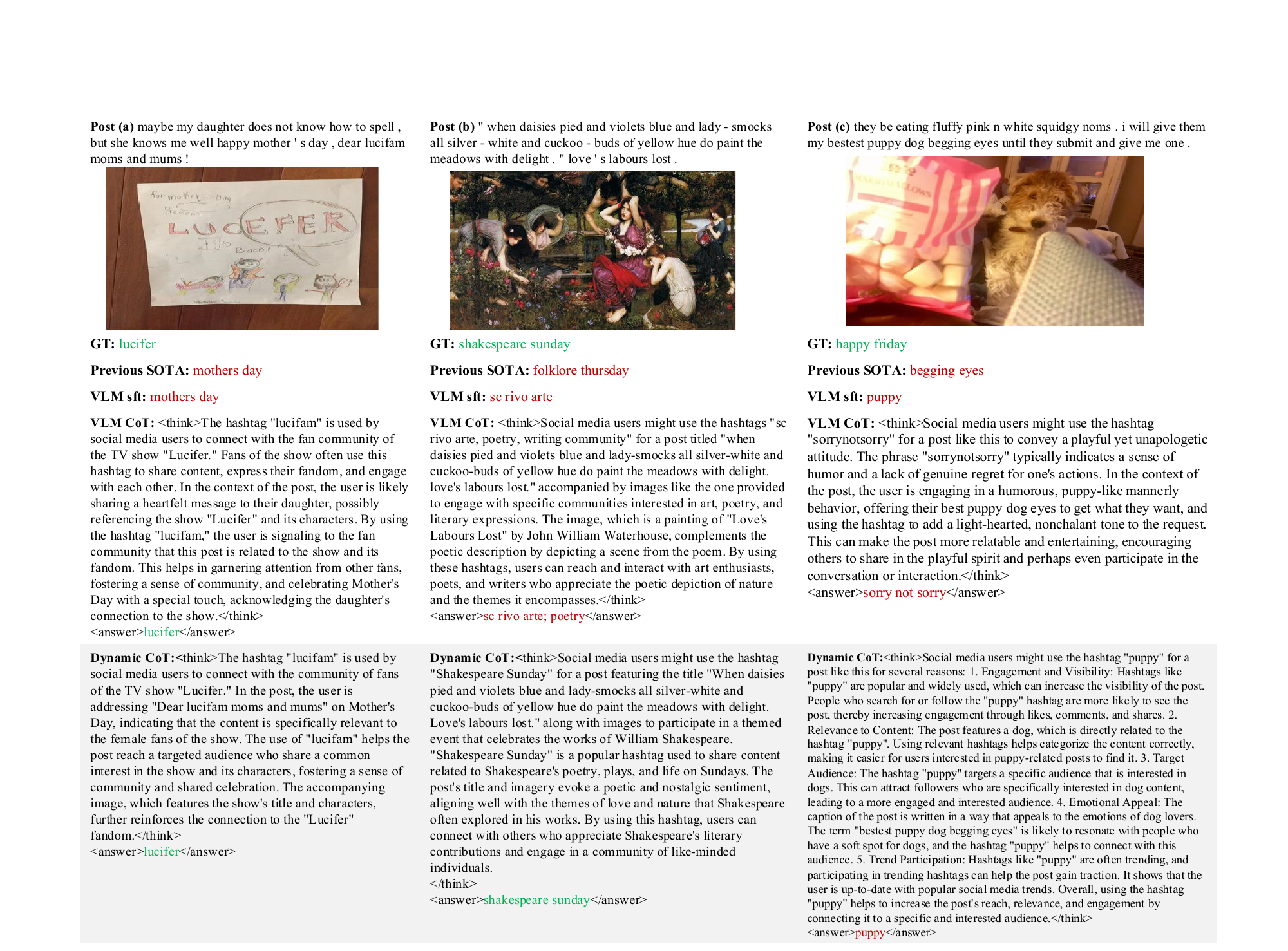}
    \caption{Detailed CoT by our proposed method.}
    \label{fig:detail_cot}
    \vspace{-2em}
\end{figure*}
Figure~\ref{fig:overfit} visualizes prediction of the SFT model on the test set, revealing its over-reliance on similarity-based memorization. The model predominantly predicts identical labels for visually analogous samples, demonstrating limited deployment of reasoning capabilities.

\textbf{Detailed CoT.}
Figure~\ref{fig:detail_cot} illustrates the Chain-of-Thought reasoning processes in our model, demonstrating its dual analytical capabilities: multimodal content analysis (textual and visual elements of social media posts) combined with social impact prediction (assessing keyphrase potential for community consensus formation and engagement generation).

\textbf{Evaluation Metrics.}
Current approaches to multimodal keyphrase prediction predominantly rely on the F1 score, yet this metric has limitations. First, it does not adequately credit predicted keyphrases that are semantically similar to the ground-truth keyphrases. Second, because user-annotated keyphrases are subjective, different users may select different hashtags for the same post, the F1 score fails to account for such variability. Accordingly, we investigated additional evaluation methods, including LLM-as-a-judge and human evaluation.

Table~\ref{tab:appendix_compare_f1_human} presents results from both GPT-4o-based and human evaluations. The model-based and human-based evaluation methods employed the following procedure: For each input, the outputs generated by different models were ranked from 1 (best) to N (worst), with a lower score indicating a higher ranking. The evaluations were based on three main criteria: (1) Correctness—whether the generated keyphrase matches or is acceptable relative to the Ground Truth; (2) Relevance—whether the keyphrase is pertinent to the post’s content; and (3) Usefulness—whether the keyphrase, when used as a hashtag, aids in the dissemination or categorization of the post. The final score for each model was derived by averaging the ranking scores across all inputs.

As shown in the table, the ranking of model performances under both model-based and human evaluation aligns closely with the trend of the F1 metric, except for the results of GPT4o and Qwen2.5-VL-72B. For these two models, discrepancies in output format and other factors caused correct answers to be misclassified as incorrect in the F1 calculation.

Additionally, it should be noted that the human evaluation process proved to be particularly challenging. As previously discussed in weakness1, many hashtags represent memes, elements of community culture, or fleeting trends. Evaluating such content requires annotators to possess a deep and broad understanding of diverse community cultures, making this process particularly challenging. To expedite the process, we uniformly sampled 20 posts from diverse scenarios within each dataset, and each post was evaluated by four annotators. The average of their ratings was then used as the final ranking result.

Based on our evaluation results, we conclude that F1-based evaluation is a reasonable metric for the actual capability of the models to a certain extent. Furthermore, evaluation methods such as LLM-as-judge demonstrate greatly improved efficiency while producing results that closely align with those from human evaluation.

\end{document}